\documentclass{bmvc2k}
\usepackage{bbm}
\usepackage{hyperref}
\usepackage{caption}

\title{Weakly Supervised Semantic Segmentation Based on Web Image Co-segmentation}

\addauthor{Tong Shen}{tong.shen@adelaide.edu.au}{1}
\addauthor{Guosheng Lin}{guosheng.lin@gmail.com}{2}
\addauthor{Lingqiao Liu}{lingqiao.liu@adelaide.edu.au}{1}
\addauthor{Chunhua Shen}{chunhua.shen@adelaide.edu.au}{1}
\addauthor{Ian Reid}{ian.reid@adelaide.edu.au}{1}

\addinstitution{
	School of Computer Science,\\
	The University of Adelaide,\\
	Australia
}
\addinstitution{
	School of Computer Science and Engineering,\\
	Nanyang Technological University,\\
	Singapore
}

\runninghead{SHEN ET AL.}{Weakly Supervised Semantic Segmentation}

\def\eg{\emph{e.g}\bmvaOneDot}

\def\etal{\emph{et al}\bmvaOneDot}
\def\etc{\emph{etc}\bmvaOneDot}
\begin{document}
	
	\maketitle
	\begin{abstract}
		Training a Fully Convolutional Network (FCN) for semantic segmentation requires a large number of masks with pixel level labelling, which involves a large amount of human labour and time for annotation. In contrast, web images and their image-level labels are much easier and cheaper to obtain. In this work, we propose a novel method for weakly supervised semantic segmentation with only image-level labels. The method utilizes the internet to retrieve a large number of images and uses a large scale co-segmentation framework to generate masks for the retrieved images. We first retrieve images from search engines, \eg Flickr and Google, using semantic class names as queries, \eg class names in the dataset PASCAL VOC 2012. We then use high quality masks produced by co-segmentation on the retrieved images as well as the target dataset images with image level labels to train segmentation networks. We obtain an IoU score of 56.9 on test set of PASCAL VOC 2012, which reaches the state-of-the-art performance.
	\end{abstract}

	\section{Introduction}
Deep Convolutional Neural Networks (DCNN) have been proven very useful in various tasks in Computer Vision including classification \cite{Krizhevsky2012, Arge2015, Technologii2013}, detection \cite{Ren2015a, Liu2015c, Redmon2015}, and semantic segmentation \cite{Chen2016, Lin2016, Lin2016a, Lin2015}. Unlike classification where the ground truth is simply image-level labels, dense prediction tasks such as semantic segmentation require pixel-level labels, which involve great annotation effort. As pointed in \cite{Bearman2015}, it takes averagely 239.7 seconds to process one image to get full supervision, and only 20 seconds to get image-level labels. Obviously, reducing the labelling time makes us achieve training data more cheaply, and further enable more training data to be collected easily.

To reduce the cost of generating pixel-level ground truth, many weakly and semi-supervised methods have been proposed. In addition to image-level labels \cite{Kolesnikova, Berthod2015}, other means of supervision have also been utilized such as bounding boxes \cite{Berthod2015}, scribble \cite{Lin2016b}, points \cite{Bearman2015} \etc. Point supervision is able to indicate the location of  objects, and bounding boxes and scribble can even indicate the location as well as the extent of objects. Image-level supervision is obviously the most challenging task where we only know the existence of objects. The objective of this work is to perform semantic segmentation only with image-level labels and try to reduce the gap between weakly supervised methods and fully supervised methods.

For weakly supervised methods with only image-level labels \cite{Bearman2015, Wei2015, Pinheiro2015}, it is very common to introduce some prior knowledge by other auxiliary methods such as Objectness \cite{Alexe2012}, Multi-scale Combinatorial Grouping (MCG) \cite{Arbelaez2014}, and saliency detection \cite{Jiang2013}. It is worth noting that these methods might rely on more than just image-level labels to be trained. For example, Objectness \cite{Alexe2012} requires bounding boxes to train and MCG requires pixel-level ground truth. Therefore, strictly speaking, using only image-level labels indicates no more supervision other than image-level labels including the training of the other methods involved.

In this work, we aim to propose a weakly supervised framework for semantic segmentation that strictly complies with the rules of image-level supervision. Our method is based on a co-segmentation method \cite{Chen}, which is an unsupervised and robust method for large scale co-segmentation. More specifically, our framework has two steps, training an initial network as a mask generator and training another network as the final model. The mask generator is trained with the retrieved images from the internet and is used to provide masks for the final model. The final model is trained by the masks from the mask generator as well as the image-level labels provided. In the first step, we first retrieve images from search engines (Flickr and Google) according to class names. Then for each class, which has a large number of images containing the same semantic object, we use co-segmentation to extract masks for each image. These masks are used to train the mask generator. In the second step, the mask generator produces masks for the target dataset, \eg PASCAL VOC 2012. With these mask as well as the image-level labels that helps to eliminate impossible predictions, we are able to train the final model with high quality ground truth. 

Our contributions are as follows:
\begin{itemize}
	\item We propose a new weakly supervised method based on co-segmentation. Apart from  image-level supervision, there is no extra supervision involved. We show that this simple two-step framework is effective for weakly supervised segmentation.
	\item We use the most popular benchmark dataset, PASCAL-VOC12 to demonstrate the performance of our framework and we achieved the state-of-the-art performance. 
\end{itemize}
	\section{Related Work}
\subsection{Weakly Supervised Semantic Segmentation}
In the literature, there have been many weakly- and semi-supervised methods proposed for semantic segmentation. Those methods utilize different forms of supervision including image-level labels \cite{Berthod2015, Kolesnikova, Oh2017, Wei2015, Pinheiro2015}, bounding boxes \cite{Dai2015}, points \cite{Bearman2015}, scribble \cite{Lin2016b}, or combined supervision \cite{Hong, Papandreou2015}. 

Our work only uses image-level labels. Therefore, we will discuss some works with the same supervision setting. Pinheiro \etal \cite{Pinheiro2015} treat weakly supervised segmentation as a Multiple Instance Learning (MIL) task. They claim that using image-level label for training makes the model learn to discriminate the right pixels, and using some extra smoothing priors can give good pixel labelling results. Pathak \etal \cite{Berthod2015} introduce a constrained CNN for weakly supervised training by setting a series constraints for object size, foreground, background \etc. Kolesnikov \etal \cite{Kolesnikova} propose a "seed, expand and constrain" framework where they employ localization cues from DCNN to find the object's location, use global weighted rank pooling to expand the mask, and use Conditional Random Fields (CRFs) to refine the boundary. Oh \etal \cite{Oh2017} combine localization cues and saliency to localize objects and obtain their extent so as to generate semantic masks. Wei \etal \cite{Wei2015} use a simple to complex framework for weakly supervised learning. They retrieve images from Flickr and use saliency maps as ground truth for training. Then they train the model in a "simple to complex" fashion as they train initial model, enhanced model and powerful model step by step. It is worth noting that using saliency has its own limitation because it is class-agnostic and the interesting region might not be the salient region detected. Our co-segmentation based framework will be more robust and respect the co-occurrence regions. More details will be discussed in Section \ref{sec:s_vs_co}.
\subsection{Co-segmentation}
The objective of co-segmentation is to segment similar objects from a pair of images or a group of images. Some methods are focused on small scale co-segmentation where only a small number of images are involved \cite{Asmuth1981, Joulin2010, Dai2013}. These bottom-up methods use low level features to find similarities between images and formulate optimization problems to find the co-segments. 

There are some other methods aiming at large scale co-segmentation with a large number of images presented including noisy data \cite{Faktor2013, Chen}. Faktor \etal \cite{Faktor2013} define co-segmentation as a composition problem where good co-segments can be easily composed and vice versa. The method is suitable for both large scale data and even a single image. Chen \etal \cite{Chen} propose an approach to combine top-down segmentation priors learned from visual subcategories and bottom-up cues to produce good co-segmentation from noisy web data. Our framework is based on this technique, which can provide us good training masks for retrieved images.
	\section{Method}
\begin{figure}[t]
	\centering
	\includegraphics[width=0.7\linewidth]{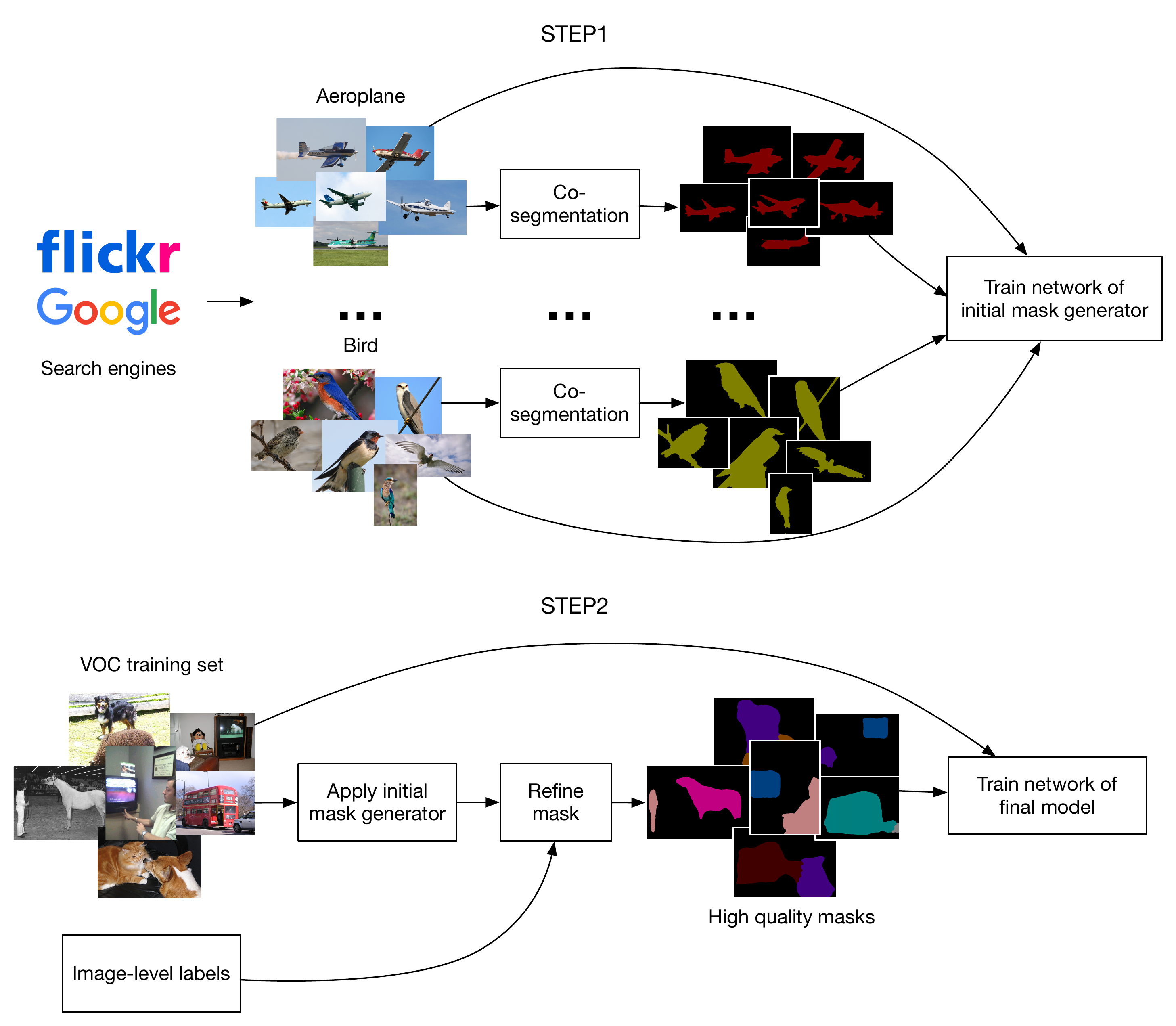}
	\caption{Illustration of the framework. In the first step, we firstly retrieve images from searching engines according to different semantic class names. Then co-segmentation is used to get pixel-wise masks for each semantic class group. Next we use these masks together with the images to train the initial mask generator. In the second step, we apply the mask generator to obtain pixel-wise masks for the target dataset, PASCAL VOC 2012, and further we refine the masks by using image-level labels to remove wrong predictions. The final model is trained with high quality masks.}
	\label{fig:pipeline}
\end{figure}
Figure \ref{fig:pipeline} shows the overall pipeline of our framework. Basically, there are two steps. In the first step, we retrieve images according to class names, \eg 20 object names in PASCAL-VOC12, and we use large scale co-segmentation method \cite{Chen} for each class group. In each group, we assume that most of the images contain the desired object in the corresponding class. Since the co-segmentation method has great ability to tolerate noise, we can control the balance between recall and precision and get high quality results. After we obtain the co-segmentation results, we treat them as ground truth masks and train a CNN as initial mask generator. In the second step, we use the mask generator to produce masks for images of the target dataset, \eg PASCAL VOC 2012. Furthermore, since we have access to image-level labels, we can use it to eliminate impossible predictions and enhance the masks. Finally we use these high quality masks as well as image-level labels to train another CNN as the final model. More details will be discussed in later sections. 

Let $\mathbf{X}$ be an image, and $\mathbf{Y}=[y_1, y_2,...,y_C]$ be the label vector, where $y_j=1$ if the image is annotated with class $j$ and otherwise $y_j=0$; $C$ is the number of class. Assume we have training data, $D=\{(\mathbf{X}_n, \mathbf{Y}_n)\}^N_{n=1}$, where $N$ is the number of training data. Our objective is to learn a dense prediction function $f(\mathbf{X};\mathbf{\theta})$ parameterized by $\mathbf{\theta}$, which models class probability of all pixels of the image.

\subsection{Saliency vs Co-segmentation}
\label{sec:s_vs_co}
In this section, let us first discuss the advantages of co-segmentation methods in terms of estimating the mask of retrieved images compared with saliency detection. In \cite{Wei2015}, saliency detection method DRFI \cite{Jiang2013} is used to generate masks for internet retrieved images. Generally speaking, saliency detection frameworks require relatively clean background and high contrast between foreground and background. For relatively complex images, saliency has its own limitation, which may detect undesired objects. One situation could be that the salient part is not the target object. Another situation could be that saliency only concentrates on the most salient part so that the other parts of the object are ignored.

In contrast, co-segmentation has more advantages in terms of localizing the desire object and finding the accurate contour. In 
\cite{Chen}, to perform co-segmentation on a group of images, \eg a group of car images, it first uses low-level features to find aligned homogeneous clusters and learns a set of visual subcategories. Then top-down segmentation priors can be created and used to extract the final segment for each image with the help of graph-cut algorithm. This bottom-up and top-down procedure can better localize the common objects in the image collection and obtain accurate contour.

\begin{figure}[t]
	\centering
	\includegraphics[width=0.7\linewidth]{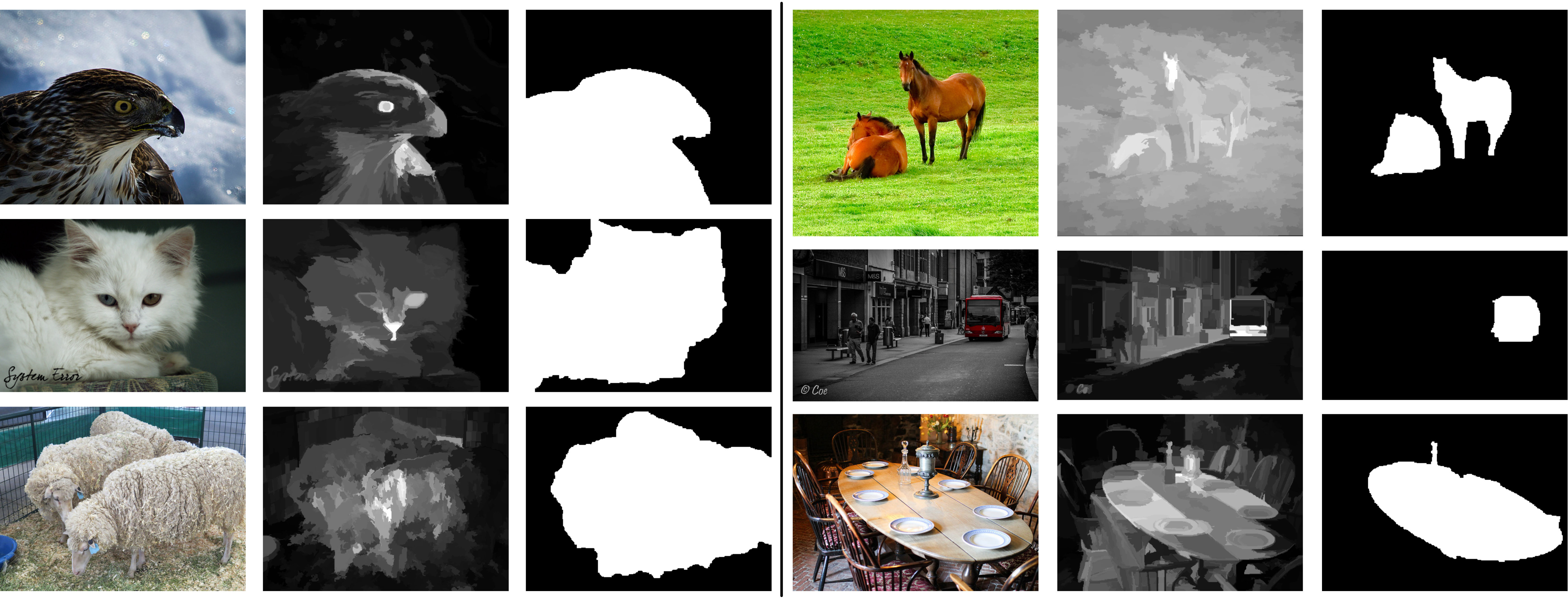}
	\caption{Comparison between saliency and co-segmentation. In each half, the first column is the original image, the middle is the saliency and the last is the segmentation mask. In the left half, these are examples when saliency fails to detect the object as a whole part. In the right half, it shows some cases where saliency gives high probability to many parts and fails to detect desired object.}
	\label{fig:saliency}
\end{figure}

Figure \ref{fig:saliency} shows some examples of saliency map produced by DRFI and co-segmentation mask. In each half, the first column is the original image; the middle is the saliency map and the last is the co-segmentation mask. In the left half, we show some example when saliency fails to detect the object as a whole part. Instead only a small portion has high probability, which makes it difficult to capture the entire object. In comparison, co-segmentation can capture the whole part. In the right half, there are some circumstances where saliency produces lots of false positives and fails to detect the part we expect. Co-segmentation can find better patterns from the large collection of images containing the common objects, which gives us more promising results.

\subsection{Training Initial Mask Generator}
To train the initial mask generator, we retrieve images from Flickr and Google and use co-segmentation to obtain estimated ground truth for each image. We define $\mathbf{M}$ as the mask, and this data set is expressed as $D_1=\{(\mathbf{X}_n, \mathbf{M}_n)\}^{N_1}_{n=1}$, where $N_1$ is the number of retrieved images. We train the initial Mask generator as in standard fully-supervised framework using softmax loss. For a single image, the loss is defined as:
\vspace{-7pt}
\begin{equation}
\mathcal{L}_1=\frac{1}{I}\sum_{i}^{I}\sum_{j}^{C}\mathbbm{1}(m_{i}=j)\log(\frac{\exp(f_{ij})}{\sum_{k}\exp(f_{ik})}),
\label{eq:softmax}
\vspace{-7pt}
\end{equation}

where $f_{ij}$ is the prediction from the network for class $j$ at spatial position $i$; $m_i$ is the training mask for spatial position $i$; $I$ is the number of spatial points.

The network structure we use is Resnet50 \cite{Technologii2013} with dilations, similar to DeepLab \cite{Chen2016}. 

\subsection{Training Final Model}
\label{sec:final_model}
After we trained the initial mask generator, we can use it to generate pixel-level labels for images of the target dataset. However, these labels could be noisy. For example, "cow" pixels could be predicted to "sheep" pixels but the image does not have sheep. Fortunately since image-level labels are accessible to these images, we can use it as constraints to eliminate impossible predictions to obtain high quality masks. For an image in the target dataset, we apply the following operation:
\vspace{-7pt}
\begin{equation}
\label{eq:generation}
m_{i}=\underset{j\in\{1,..,C\}}{\arg\max}y_{j}f_{ij},
\vspace{-7pt}
\end{equation}
where $y_{j}=1$ if class $j$ is shown and otherwise $0$; $f_{ij}$ is the score for class $j$ at position $i$; $m_i$ is the label for position $i$.

So far, we have the masks for all the images in the target dataset. The training set can be expressed as $D=\{(\mathbf{X}_n, \mathbf{Y}_n, \mathbf{M}_n)\}^N_{n=1}$. 

We train the final model also in fully supervised setting. Furthermore, in order to include more context information, we borrow the idea from \cite{Shen2017} where multi-label classification is combined to enforce scene consistency. For simplicity, we only add one branch for global context. Thus the training process is guided by two losses, softmax loss and multi-label loss. The softmax loss is the same as in Equation \ref*{eq:softmax}, while the multi-label loss for single image is a binary logistic loss expressed as:
\vspace{-7pt}
\begin{equation}
\mathcal{L}_{2}=\frac{1}{C}\sum_{j}y_{j}\log(\frac{1}{1+e^{-p_{j}}})+(1-y_{j})\log(\frac{e^{-p_{j}}}{1+e^{-p_{j}}}),
\vspace{-7pt}
\end{equation}
where $p_j$ is the prediction for class $j$.

The objective is to minimize the combined loss for all images:
\vspace{-7pt}
\begin{equation}
\mathcal{L} = \sum_{n}^{N}\mathcal{L}_1^n + \lambda\mathcal{L}_2^n,
\label{eq:loss}
\vspace{-7pt}
\end{equation}
where $\lambda$ controls balance between losses.

Intuitively, multi-label classification imposes constraints to the segmentation prediction so that the prediction would be consistent to the scene globally. Implausible predictions will be largely modified. The final model still uses Resnet50 with dilations as basic structure. The difference is the multi-label branch. The branch shares some low-level layers with the main branch and has its own high level layers.
	\section{Experiments}
\subsection{Datasets}
\textbf{Retrieved Dataset:} We construct a dataset by retrieving images from Flickr and Google. By using the corresponding class names in our target dataset PASCAL VOC 2012 as queries, we retrieve 75800 images in total\footnote{Can be found at \href{https://ascust.github.io/WSS/}{https://ascust.github.io/WSS/}}. It is worth noting that we do not have any filtering process as in \cite{Wei2015} where they adopt multiple methods to filter the crawled images. Since the co-segmentation we use has good tolerance to noise, we simply use all the images crawled. The only preprocessing is resizing all the images so that the maximum dimension is 340. This dataset is used to train our initial mask generator.

\textbf{PASCAL VOC 2012:} We train and evaluate the final model on this dataset. The original dataset \cite{Everingham2010} contains 1464 training images, 1449 validation images and 1456 testing images. The dataset is further augmented to 10582 training images as in \cite{Hariharan}. There are totally 21 semantic classes in the benchmark. The evaluation metric is the standard Intersection over Union (IoU) averaged on all 21 classes.

\subsection{Experiment Setup}
\textbf{Co-segmentation:} We use the public available code of \cite{Chen}. For all the retrieved images, the images belonging to the same semantic class will be in a large group for co-segmentation. Therefore, there are 20 groups (excluding background class) of images. We used all default settings without tweaking any parameters. After obtaining the results, we only keep the images that have foreground pixels between 20\%-80\% of the whole image, which results in 37211 images.

\textbf{Initial mask generator:} We use the images from co-segmentation to train the mask generator. To make the images compatible to PASCAL VOC 2012, we resize the original images and masks so that the maximum dimension is 500. The network is trained with public toolbox MXNet \cite{Chen2015b}. The network structure is basically Resnet50 with dilated convolutions, which has resolution of 1/8, similar to Deeplab structure \cite{Chen2016}. We use standard Stochastic Gradient Descent (SGD) with batch size 16, crop size 320, learning rate 16e-4, weight decay 5e-4 and momentum 0.9. The training takes around 5000 iterations. The learning rate is decreased by factor of 10 once for further fine-tuning. It is worth noting that longer training time will worsen the performance because the model might fit to some noisy data or bad quality data. 

\textbf{Final model:} After having the mask generator trained, we use it to generate masks for 10582 training images in PASCAl VOC 2012. Following Equation \ref{eq:generation}, we make predictions with the image-level label constraint. Also we use multi-scale inference to combine results at different scales to increase the performance, which is common practice as in \cite{Lin2016a} \cite{Cvpr2017}. The final model is trained only using these data. The network structure is a little different from the initial mask generator. As mentioned in Section \ref{sec:final_model}, we include a global multi-label branch to enforce the scene consistency globally, it turns out to be very useful, as shown in later ablation study in Section \ref{sec:ablation}. The parameters are similar to the training process for the mask generator. The extra parameter $\lambda$ in Equation \ref{eq:loss} is set to 1.0. Besides, we train the model for 11000 iterations and further fine-tune it for several iterations with the learning rate decreased by factor of 10. The final outputs are post-processed by CRF\cite{Krahenbuhl2012}.

\subsection{Experimental Results}
\begin{table}[t]
	\begin{centering}
	\scalebox{0.6}{
	\begin{tabular}{|c||c|c|c|c||c|c|c|c|c|}
		\hline 
		 & \multicolumn{4}{c||}{Validation set} & \multicolumn{5}{c|}{Test set}\tabularnewline
		\cline{2-10} 
		& \cite{Berthod2015} & \cite{Kolesnikova} & \cite{Wei2015} & Ours & \cite{Berthod2015} & \cite{Kolesnikova} & \cite{Papandreou2015} & \cite{Wei2015} & Ours\tabularnewline
		\hline 
		background & 68.5 & 82.4 & 84.5 & \textbf{85.8} & 71 & 83.5 & 76.3 & 85.2 & \textbf{86.9}\tabularnewline
		aeroplane & 25.5 & 62.9 & \textbf{68} & 53.0 & 24.2 & 56.4 & 37.1 & \textbf{62.7} & 57.9\tabularnewline
		bike & 18.0 & \textbf{26.4} & 19.5 & 24.0 & 19.9 & \textbf{28.5} & 21.9 & 21.1 & 26.3\tabularnewline
		bird & 25.4 & 61.6 & 60.5 & \textbf{69.4} & 26.3 & 64.1 & 41.6 & 58 & \textbf{65.1}\tabularnewline
		boat & 20.2 & 27.6 & \textbf{42.5} & 36.7 & 18.6 & 23.6 & 26.1 & \textbf{31.4} & 28.3\tabularnewline
		bottle & 36.3 & 38.1 & 44.8 & \textbf{64.3} & 38.1 & 46.5 & 38.5 & 55 & \textbf{63.9}\tabularnewline
		bus & 46.8 & 66.6 & 68.4 & \textbf{81.9} & 51.7 & 70.6 & 50.8 & 68.8 & \textbf{80.6}\tabularnewline
		car & 47.1 & 62.7 & 64.0 & \textbf{64.6} & 42.9 & 58.5 & 44.9 & 63.9 & \textbf{70.7}\tabularnewline
		cat & 48.0 & \textbf{75.2} & 64.8 & 74.5 & 48.2 & \textbf{71.3} & 48.9 & 63.7 & 68.8\tabularnewline
		chair & 15.8 & \textbf{22.1} & 14.5 & 11.4 & 15.6 & \textbf{23.2} & 16.7 & 14.2 & 15.5\tabularnewline
		cow & 37.9 & 53.5 & 52.0 & \textbf{70.2} & 37.2 & 54 & 40.8 & 57.6 & \textbf{67.1}\tabularnewline
		diningtable & 21.0 & 28.3 & 22.8 & \textbf{34.2} & 18.3 & 28 & 29.4 & 28.3 & \textbf{37.3}\tabularnewline
		dog & 44.5 & 65.8 & 58.0 & \textbf{72.7} & 43 & 68.1 & 47.1 & 63 & \textbf{74.2}\tabularnewline
		horse & 34.5 & 57.8 & 55.3 & \textbf{66.3} & 38.2 & 62.1 & 45.8 & 59.8 & \textbf{70.1}\tabularnewline
		motorbike & 46.2 & \textbf{62.3} & 57.8 & 60.5 & 52.2 & \textbf{70} & 54.8 & 67.6 & 69.9\tabularnewline
		person & 40.7 & 52.5 & \textbf{60.5} & 42.3 & 40 & 55 & 28.2 & \textbf{61.7} & 45.9\tabularnewline
		plant & 30.4 & 32.5 & 40.6 & \textbf{45.9} & 33.8 & 38.4 & 30 & 42.9 & \textbf{50.6}\tabularnewline
		sheep & 36.3 & 62.6 & 56.7 & \textbf{71.6} & 36 & 58 & 44 & 61 & \textbf{68.0}\tabularnewline
		sofa & 22.2 & 32.1 & 23.0 & \textbf{34.7} & 21.6 & 39.9 & 29.2 & 23.2 & \textbf{43.9}\tabularnewline
		train & 38.8 & 45.4 & 57.1 & \textbf{66.6} & 33.4 & 38.4 & 34.3 & 52.4 & \textbf{58.7}\tabularnewline
		tv/monitor & 36.9 & 45.3 & 31.2 & \textbf{53.3} & 38.3 & \textbf{48.3} & 46 & 33.1 & 45.4\tabularnewline
		\hline 
		mean & 35.3 & 50.7 & 49.8 & \textbf{56.4} & 35.6 & 51.7 & 39.6 & 51.2 & \textbf{56.9}\tabularnewline
		\hline 
	\end{tabular}}
	\par\end{centering}
	\caption{Result of different image-level semantic segmentation methods on validation and test set of PASCAL VOC 2012.}
	\label{tab:results}
\end{table}

Table \ref{tab:results} shows the IoU scores of our method and other weakly-supervised methods on validation set and test set of PASCAL VOC 2012, where we achieve IoU of 56.9\footnote{http://host.robots.ox.ac.uk:8080/anonymous/NNRJCF.html} on test set. From the table, we can easily see that our method outperforms other methods in majority of classes. For some classes, our method increases the score by a big margin, for example, bottle, cow, bus. This big improvement benefits from the good quality masks produced by co-segmentation. For some low score classes, such as chair and person, we suspect it is due to the extreme co-occurrence, for which co-segmentation always treats them as a whole part. For instance, chair always shows with dining-table or motorbike always appears with person. Some examples can be seen in later failure analysis in Section \ref{sec:failure}. 

\begin{figure}[t]
	\centering
	\includegraphics[width=0.7\linewidth]{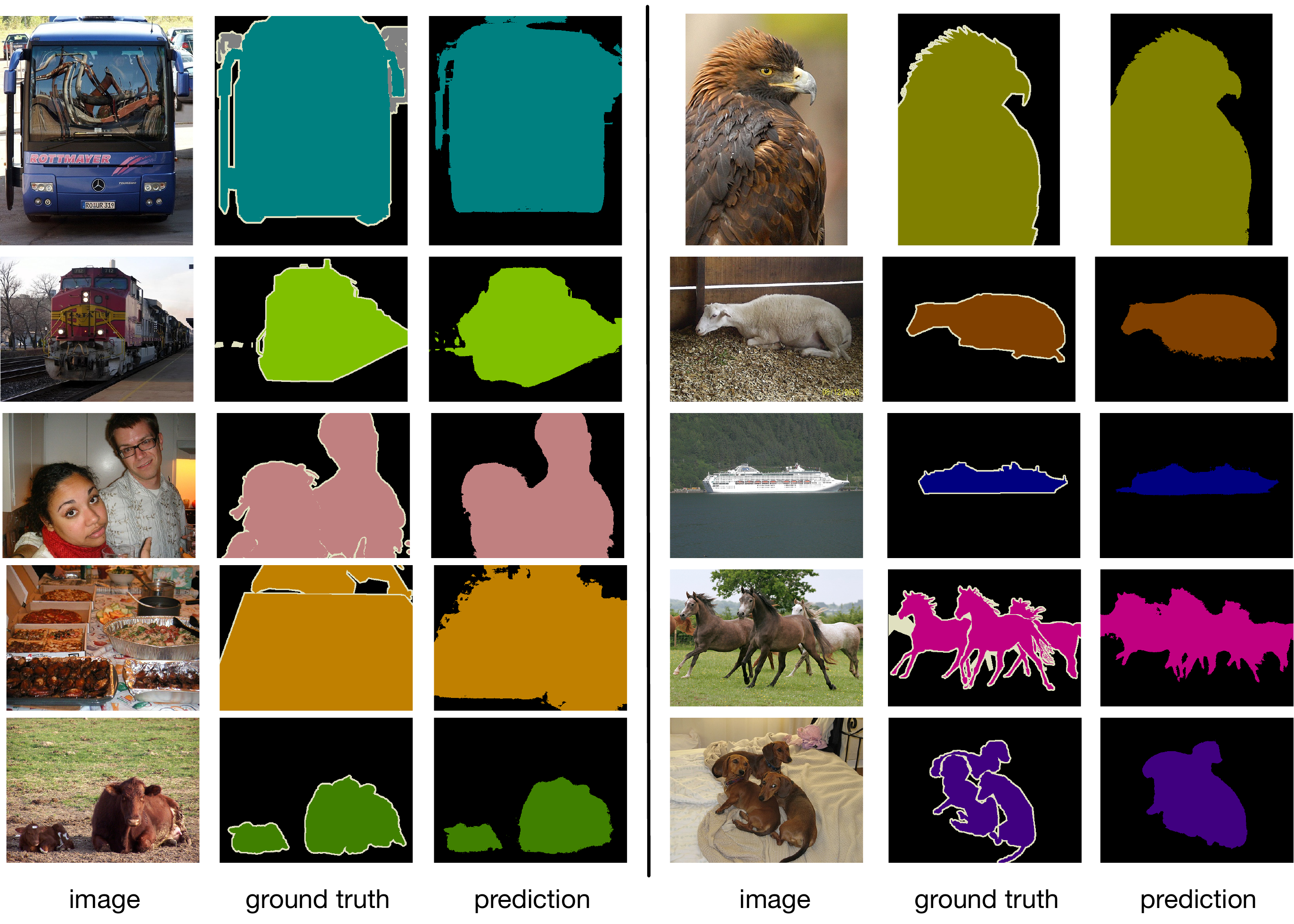}
	\caption{Qualitative segmentation results on PASCAL VOC 2012 validation set.}
	\label{fig:examples}
\end{figure}

We also compare the results with other methods with stronger supervision, as shown Table \ref{tab:other_methods}. In the upper half, those are methods that either use stronger supervision apart from image-level labels or use other techniques that involve other supervisions. In \cite{Lin2016b} and \cite{Dai2015}, their supervision, scribble and bounding box can indicate not only the location of objects but also the extent of objects, which is far more informative than image-level labels. In \cite{Hong}, they use pixel-level masks, which, even with relatively small number, can greatly boost the performance. For \cite{Oh2017} and \cite{B2016}, the techniques involved, saliency and MCG, require additional supervision, \eg bounding boxes or pixel-level masks to train. We can safely conclude that with stronger supervision, the performance gets better accordingly. Our method only uses the weakest supervision but can reach competitive performance compared with other supervision settings. Some qualitative results are shown in Figure \ref{fig:examples}.

\begin{table}[t]
	\begin{centering}
		\scalebox{0.6}{
		\begin{tabular}{|c|c|c|c|}
			\hline 
			Method & Val & Test & Supervision\tabularnewline
			\hline 
			\cite{Lin2016b} & 63.1 & - & Scribble supervision\tabularnewline
			\cite{Dai2015} & 62.0 & 64.6 & Bounding box supervision + MCG\tabularnewline
			\cite{Hong} & 62.1 & 62.5 & 20 fully supervised images for each class\tabularnewline
			\cite{Oh2017} & 55.7 & 56.7 & Extra bounding box supervision for saliency detection\tabularnewline
			\cite{B2016} & 54.3 & 55.5 & MCG\tabularnewline
			\cite{Bearman2015} & 46.1 & - & Point supervision on each class\tabularnewline
			\hline 
			\cite{Berthod2015} & 35.3 & 35.6 & \tabularnewline
			\cite{Kolesnikova} & 50.7 & 51.7 & \tabularnewline
			\cite{Papandreou2015} & - & 39.6 & Image-level labels (weakest)\tabularnewline
			\cite{Wei2015} & 49.8 & 51.2 & \tabularnewline
			Ours & 56.4 & 56.9 & \tabularnewline
			\hline 
	\end{tabular}}
		\par\end{centering}
	\caption{Comparison with methods with stronger supervision. From the table, it can be safely concluded that generally stronger supervision leads to better performance. Although our method only uses imagel-level labels, which is the weakest supervision, we have achieved very competitive results. }
	\label{tab:other_methods}
	
\end{table}

\subsection{Ablation Study}
\label{sec:ablation}
\begin{table}[t]
	\begin{centering}
		\scalebox{0.6}{
		\begin{tabular}{|c|c|}
			\hline 
			Model & IoU\tabularnewline
			\hline 
			\hline 
			Initial mask generator & 48.3\tabularnewline
			\hline 
			Simple final model & 53.3\tabularnewline
			\hline 
			Final model with multi-label module & 55.1\tabularnewline
			\hline 
			Final model+MS infer+CRF & \textbf{56.4} \tabularnewline
			\hline 
		\end{tabular}}
		\par\end{centering}
	\caption{Results with different settings on validation set.}
	\label{tab:ablation}
\end{table}

To analyse the effect of each part in the framework, we conducted an ablation experiment, whose results on validation set are shown in Table \ref{tab:ablation}. In the first stage of the framework, the initial mask generator is trained to reach IoU of 48.3. Then if we use the same structure without the global multi-label module mentioned in Section \ref{sec:final_model}, we only get IoU of 53.3. With the help of the global multi-label module, this performance can be increased to 55.1. With some post-processing, multi-scale inference and CRF, we get 56.4 on validation set.

\subsection{Failure Analysis}
\label{sec:failure}
\begin{figure}[t]
	\centering
	\includegraphics[width=0.7\linewidth]{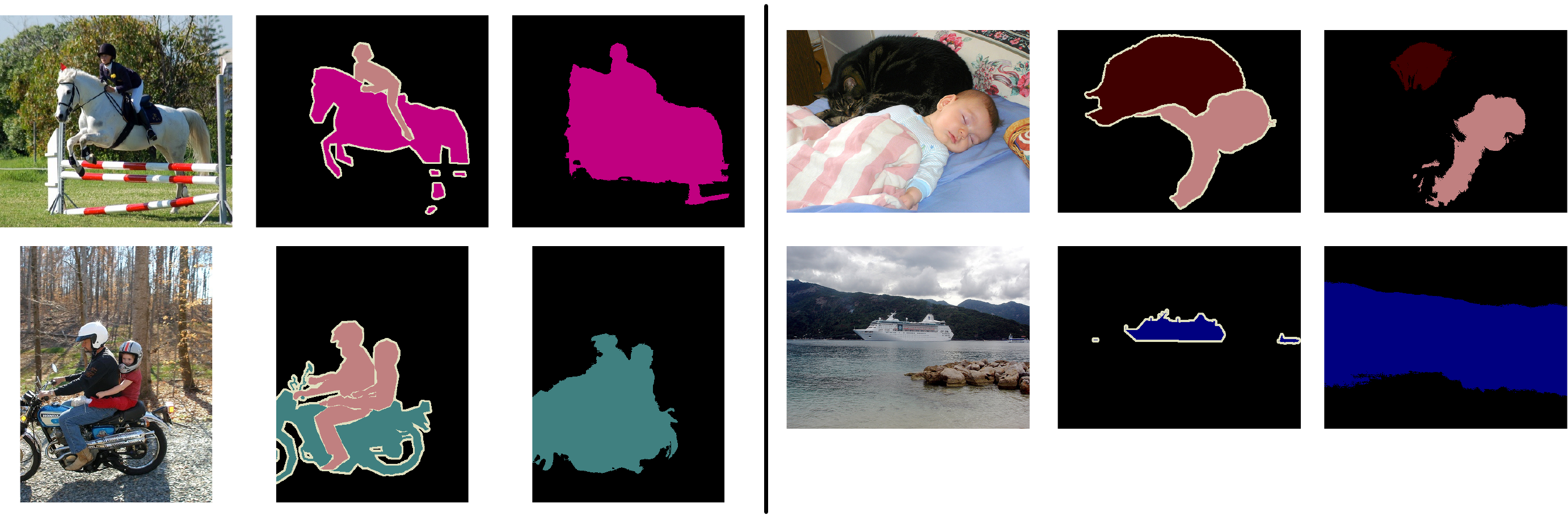}
	\caption{Examples of some failure cases.}
	\label{fig:failure}
\end{figure}

In Figure \ref{fig:failure}, there are some failure cases presented. In the left half, it is a type of failure where two objects are recognized as one part due to strong co-occurrence. The right half shows another type of failure where the object is underestimated or overestimated. The reason could be some bad quality masks produced by co-segmentation.

	\section{Conclusion}
We have presented a framework for weakly supervised semantic segmentation using only image-level labels. The framework utilises co-segmentation and retrieved images from the internet to obtain training data with pixel-level masks. Our two-step framework uses these high quality masks as well as the image-level labels of the target dataset to train a semantic segmentation network. Based on the experiments shown on a popular benchmark dataset, we show that our simple but effective framework reaches the state-of-the-art performance.

	\section*{Acknowledgements}
	This research was supported by the Australian Research Council through the Australian Centre for Robotic Vision (CE140100016). C. Shen's participation was supported by an ARC Future Fellowship (FT120100969). I. Reid's participation was supported by an ARC Laureate Fellowship (FL130100102).
	\clearpage
	\bibliography{egbib}
\end{document}